\documentclass[runningheads]{llncs}
\usepackage[T1]{fontenc}
\usepackage{booktabs}
\usepackage{multirow}
\usepackage{graphicx}
\usepackage{amsmath}
\usepackage{amssymb}
\usepackage{graphicx,verbatim}
\begin{document}
\title{
Learning Where to Look: A Reinforcement Learning Framework for Robust Micro-Ultrasound Prostate Cancer Detection
}
%
\titlerunning{Learning Where to Look}

\author{
Mohammad Mahdi Abootorabi\inst{1,3} \and
Sina Namazi\inst{1} \and
Armin Saadat\inst{1} \and
Lyuyang Wang\inst{1} \and
Obed Dzikunu\inst{1,3} \and
Paul F. R. Wilson\inst{2,3} \and
Zhuoxin Guo\inst{2,3} \and
Brian Wodlinger\inst{4} \and
Parvin Mousavi\inst{2,3} \and
Purang Abolmaesumi\inst{1}
}
\index{Abootorabi, Mohammad Mahdi}
\index{Namazi, Sina}
\index{Saadat, Armin}
\index{Wang, Lyuyang}
\index{Dzikunu, Obed}
\index{Wilson, Paul F. R.}
\index{Zhuoxin, Guo}
\index{Wodlinger, Brian}
\index{Mousavi, Parvin}
\index{Abolmaesumi, Purang}

\authorrunning{M. M. Abootorabi et al.}

\institute{
The University of British Columbia, Vancouver, BC, Canada \\
\email{\{mahdi.abootorabi,purang\}@ece.ubc.ca}
\and
Queen's University, Kingston, ON, Canada
\and
Vector Institute, Toronto, ON, Canada
\and
Exact Imaging, Markham, ON, Canada
}

\maketitle 
\vspace{-6mm}
\begin{abstract}
Micro-ultrasound ($\mu$US) is a new, emerging, and promising imaging modality for prostate cancer (PCa) detection, but accurate identification of suspicious tissue remains highly dependent on clinical experience, leading to substantial inter-observer variability. Machine-learning assistance can reduce this variability; however, training reliable deep models is challenging because supervision is sparse and noisy—typically limited to core-level histopathology outcomes (e.g., cancer grade and its percentage in a biopsy core) without pixel-level lesion annotations and under severe class imbalance. We introduce Prost-RL, which reframes $\mu$US PCa detection as a spatially aware, policy-driven inference problem by learning where to look before decoding. Prost-RL integrates a lightweight reinforcement-learning policy into a foundation-model encoder–decoder to generate interpretable spatial attention maps that act as soft prompts for both cancer-likelihood heatmap prediction and image-level classification. We further propose Adaptive Policy Optimization (APO) to stabilize hybrid supervised–RL training and a noise-robust objective combining symmetric cross-entropy with negative-entropy regularization to mitigate weak-label noise and encourage sharp localization. On a cohort of 6{,}607 biopsy cores from 693 patients across five clinical sites, Prost-RL achieves $79.0\pm3.5$ AUROC with $64.6\pm6.3$\% sensitivity at 80\% specificity for core-level detection (+2.1 AUROC and +4.5 sensitivity points over the strongest baseline), and $79.3\pm5.8$ AUROC for clinically significant cancer classification. The learned policy highlights biopsy-aligned regions, providing transparent, spatially grounded evidence alongside quantitative risk predictions. Code is available at: \url{https://github.com/DeepRCL/Prost-RL}.


\end{abstract}

\section{Introduction}
Prostate cancer (PCa) remains a leading cause of cancer mortality in men \cite{bray2024global}. While multi-parametric MRI (mpMRI) improves risk stratification \cite{reisaeter2018optimising,gavade2023automated}, its cost and infrastructure requirements limit broad access \cite{hutchinson2017cost}, motivating biopsy-targeting workflows based on ultrasound alone. High-resolution micro-ultrasound ($\mu$US) is an emerging and promising modality for PCa detection, operating at frequencies up to 29\,MHz to resolve prostate micro-architecture with MRI-comparable accuracy \cite{dubois2025micro,klotz2020comparison}. However, interpretation of $\mu$US is highly dependent on clinical experience: speckle noise, artifacts, and heterogeneous tumor appearance make suspicious regions subtle and variable \cite{harmanani2025trusworthy,gilany2022towards}, leading to substantial inter-observer variability. Automating $\mu$US interpretation with machine learning can help standardize targeting and reduce observer variability.

A central barrier is supervision. In routine biopsy, pathology is available only at the \emph{core level}: each acquired core provides a global cancer label, grade, and involvement percentage, but no pixel-level lesion delineation. Supervision is sparse (restricted to the needle region), noisy (small tumors may be missed), and strongly imbalanced (benign tissue dominates), making it difficult to train deep models that localize cancer and predict clinically significant disease reliably.

Recent work has progressed from patch-based CNNs to foundation-model (FM) adaptations that mitigate weak labels and domain shift via self-supervision, multiple-instance learning (MIL), temporal/radio-frequency modeling, prompting with clinical priors, and knowledge distillation \cite{elghareb2025proteus,harmanani2025cinepro,harmanani2025trusworthy,willis2026guideusgradeinformedunpaireddistillation,wilson2025prostnfoundprospectivestudyusing,wilson2024prostnfound}. Yet, most approaches remain \emph{passive}: they process the full image holistically and rely on implicit attention to discover lesions. This becomes brittle under weak and noisy labels, where fine-tuning large decoders can amplify spurious regions and degrade localization \cite{han2025region,wang2025weakmedsam}. These observations motivate an explicit mechanism that \emph{learns where to look} before decoding.

In this paper, we introduce \textbf{Prost-RL}, a policy-driven framework that integrates a lightweight reinforcement-learning (RL) policy into an FM encoder-decoder to produce interpretable spatial attention maps serving as soft prompts for heatmap localization and image-level classification. Prost-RL is trained end-to-end with \textbf{Adaptive Policy Optimization (APO)}, a hybrid supervised-RL objective inspired by alignment methods, and with noise-robust loss (symmetric cross-entropy and negative-entropy regularization) to stabilize learning from sparse, weak core-level labels. On a multi-center retrospective $\mu$US cohort of 6{,}607 biopsy cores from 693 patients across five sites, Prost-RL improves core-level detection to $79.0\pm3.5$ AUROC with $64.6\pm6.3$\% sensitivity at 80\% specificity (+2.1 AUROC and +4.5 sensitivity points over the strongest baseline) and achieves $79.3\pm5.8$ AUROC for clinically significant cancer classification, while providing biopsy-aligned attention maps for transparent decision support.

\begin{figure}[t]
\includegraphics[width=\textwidth]{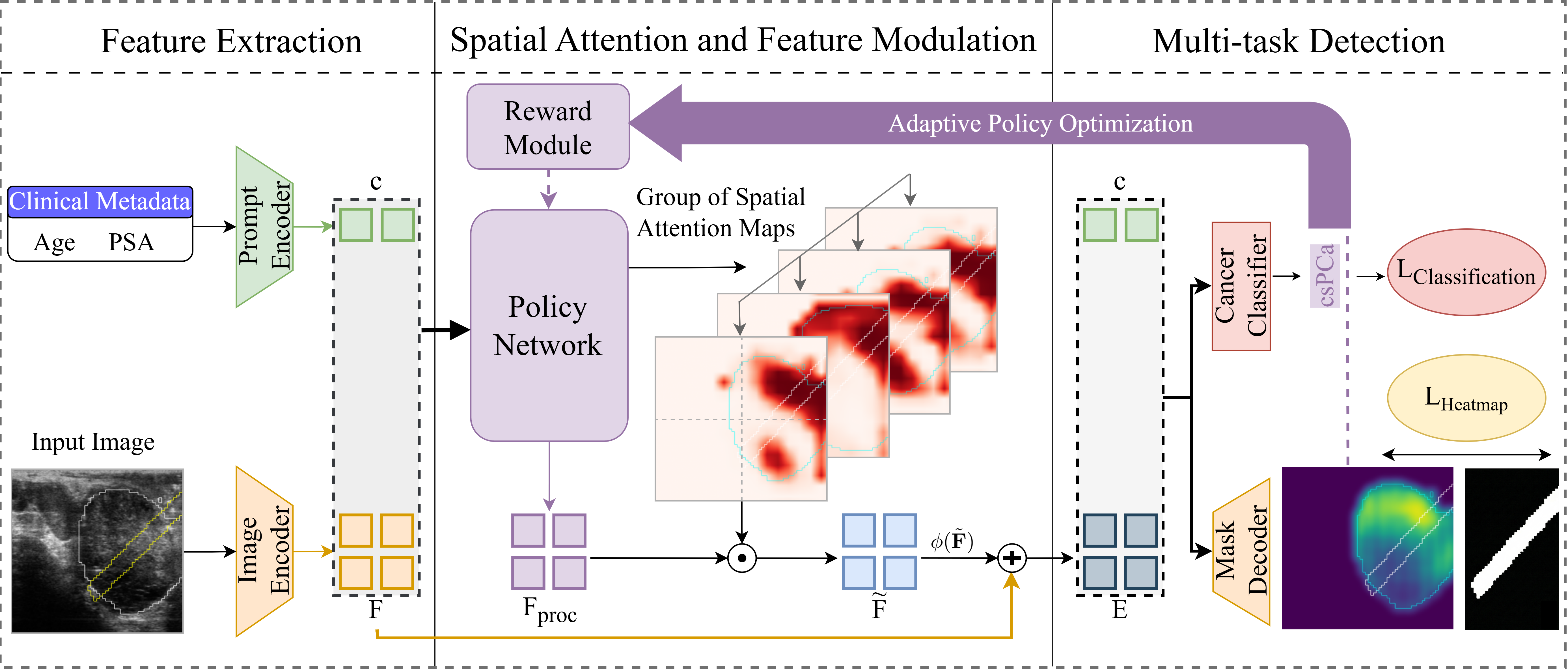}
\caption{Prost-RL overview. Image and clinical prompt encoders produce features $F$ and embedding $c$. A spatial policy $\pi_\theta$ generates an attention map $\alpha$ that modulates $F$ into shared embeddings $E$, decoded by a heatmap head and a csPCa classifier.}
\label{fig1}
\end{figure}

\section{Materials and Method}
\subsection{Data}\label{sub:data}

We utilized a private multi-center $\mu$US dataset of 6,607 biopsy cores from 693 patients, collected prospectively across five clinical sites (2013–2016) under an approved trial (ClinicalTrials.gov: NCT02079025 \cite{rohrbach2018high}) with informed consent. Biopsies were performed using the ExactVu $\mu$US system (Exact Imaging, Markham, Canada), acquiring B-mode sagittal-plane images (depth 28~mm, width 46.06~mm) at 10–12 cores per patient; the frame immediately preceding needle firing was extracted per core. Clinical metadata included patient age, PSA, PSA density (PSAD), and anatomical core location (POS) as 2D base-apex and medial-lateral coordinates. Core-level histopathology provided ground-truth ISUP Grade Group (GG) and proportional cancer involvement, assigned to annotated needle trace regions as weak spatial supervision. Positive cores (GG~$\geq$~1) comprised clinically insignificant (isPCa, GG=1–2; $n=480$) and clinically significant (csPCa, GG~$\geq$~3; $n=400$) cases, yielding 880 positives against 5,727 benign cores (86\% benign). Biopsy needle trace and whole-gland binary masks were provided alongside images. Evaluation followed patient-level five-fold cross-validation with center-stratified folds \cite{wilson2024prostnfound}; benign cores from patients with any malignant core were excluded from training to reduce label ambiguity, with no involvement-threshold filtering or benign undersampling applied. Images were resized to $256\times256$ and normalized to $[0,1]$; masks to $64\times64$ to match attention map resolution.

\subsection{Method}\label{sub:method}
Fig.~\ref{fig1} provides an overview of \textit{Prost-RL}. Our design is encoder–decoder agnostic; we instantiate it on top of
ProstNFound+ \cite{wilson2025prostnfoundprospectivestudyusing}, currently the strongest baseline for $\mu$US PCa detection, because it couples a pretrained medical foundation model (MedSAM~\cite{MedSAM}) with lightweight adapter layers and a clinical prompt encoder. \textit{Prost-RL} contributes three additions to this backbone: (i)~an \emph{attention policy network} that learns \emph{where to look} before
features reach the decoder, (ii)~a noise-robust training objective to handle weak, proportional biopsy labels, and (iii)~reinforcement-learning-tuning that optimises a ranking-aware reward. Training proceeds with a supervised warm-up, followed by RL refinement.

Given an input image $\mathbf{x}$, the MedSAM encoder produces spatial features $\mathbf{F} = \mathrm{Enc}(\mathbf{x}) \in \mathbb{R}^{C\times H\times W}$. Rather than passing these raw features directly for prediction, they are first explicitly modulated by the attention policy. The resulting modulated dense embeddings, together with shared sparse prompt embeddings (encoding clinical metadata), are passed simultaneously to two distinct heads: (i) a heatmap decoder for pixel-level cancer likelihood, and (ii) an image-level classifier specifically for clinically significant prostate cancer (csPCa) detection. Joint training on these shared inputs ensures that the explicitly learned spatial focus actively supports both precise tumor localization and global risk grading.

\noindent\textbf{Spatial Attention Policy and Feature Modulation.}
Under weak proportional supervision, models tend to produce spatially diffuse predictions.
We introduce an explicit attention policy $\pi_\theta$ that learns a spatial distribution
over encoder locations. The policy processes $\mathbf{F}$ and clinical metadata $\mathbf{c}$.
Clinical features are embedded into a channel-wise gate
$\mathbf{g}\in\mathbb{R}^C$, which modulates processed features via element-wise
multiplication. A convolutional head produces spatial logits that are masked outside the
prostate region and normalized to obtain attention weights $\boldsymbol{\alpha}$. Attention-weighted features are computed and injected through residual modulation:
$
\tilde{\mathbf{F}} = \mathbf{F}_{\mathrm{proc}} \odot \boldsymbol{\alpha},
\;
\mathbf{E} = \mathbf{F} + \phi(\tilde{\mathbf{F}}),
$
where $\mathbf{F}_{\mathrm{proc}}$ denotes the policy-processed features (after
convolution and clinical gating), and $\phi(\cdot)$ is a bias-free $1\times1$
projection with GELU nonlinearity. This additive modulation ensures zero attention
yields zero contribution while preserving base features. Modulated embeddings
$\mathbf{E}$ is shared by both decoder heads, and $\boldsymbol{\alpha}$ provides an interpretable spatial map of the model's attention.

\noindent\textbf{Noise-Robust Weakly Supervised Training.}
Ground-truth is available only as a core-level involvement proportion $q\in[0,1]$ per needle trace. Pixel-level BCE forces the model to predict cancer everywhere in a positive core, producing diffuse heatmaps. ProstNFound+~\cite{wilson2025prostnfoundprospectivestudyusing} relaxes this by applying BCE against $q$ over the needle-prostate intersection, but standard CE fully trusts noisy labels: the gradient $\partial\mathcal{L}/\partial\hat{p} = -q/\hat{p} + (1-q)/(1-\hat{p})$ diverges when the model confidently disagrees, causing memorization of erroneous proportions.

We introduce a Symmetric Cross-Entropy (SCE)~\cite{wang2019symmetric} objective combined with entropy minimization as our heatmap loss. SCE augments CE with a Reverse Cross-Entropy (RCE) term, $\mathcal{L}_{\mathrm{SCE}} = \alpha\,\mathcal{L}_{\mathrm{CE}}(\hat{p},q) +
\beta\,\mathcal{L}_{\mathrm{RCE}}(\hat{p},q)$,
where $\mathcal{L}_{\mathrm{RCE}}(\hat{p},q) = -\hat{p}\log q_{\epsilon}
-(1{-}\hat{p})\log(1{-}q_{\epsilon})$ and $q_\epsilon=\mathrm{clip}(q,\epsilon,1{-}\epsilon)$.
Because the RCE gradient with respect to $\hat{p}$ is $-\log q_\epsilon + \log(1{-}q_\epsilon)$, a constant independent of the prediction, the corrective signal remains bounded under strong model–label disagreement, providing implicit noise tolerance without discarding supervision. However, proportion-based losses only constrain the average prediction: a model outputting $q$ uniformly across all pixels satisfies $\mathcal{L}_{\mathrm{SCE}}$
identically to one that localizes precisely. To resolve this degeneracy, we add a pixel-level entropy regularizer,
$\mathcal{L}_{\mathrm{ent}} = \lambda_{H}|\mathcal{R}|^{-1}
\sum_{(i,j)\in\mathcal{R}}\mathcal{H}(\hat{y}_{i,j})$, where $\mathcal{R}$ is the valid spatial region defined by the intersection of the prostate and needle masks, $\hat{y}_{i,j}$ is the predicted pixel probability, and $\mathcal{H}$ is the binary entropy. By minimizing $\mathcal{L}_\mathrm{ent}$, we penalize uncertain, mid-range per-pixel predictions and drive the model toward spatially sharp, decisive heatmaps.
The total supervised objective is $\mathcal{L}_{\mathrm{sup}} = \mathcal{L}_{\mathrm{SCE}} + \mathcal{L}_{\mathrm{ent}} + \mathcal{L}_{\mathrm{clf}}$,
where $\mathcal{L}_{\mathrm{clf}}$ is a balanced CE on the classification head.

\noindent\textbf{Adaptive Policy Optimization.}
While our noise-robust supervised objective stabilizes learning from weak proportional labels, it inherently optimizes absolute prediction confidence on a strictly per-sample basis. However, clinical diagnosis heavily relies on relative risk stratification and is effectively a ranking problem. Furthermore, supervision is available only at the core level, providing no direct guidance on which spatial regions truly correspond to malignant tissue. Consequently, this supervision can lead to suboptimal spatial focus.

To refine the spatial policy and optimize this ranking capability, we introduce APO as a post–warm‑up training stage. The goal is not merely to increase confidence, but to bias the spatial policy toward attention patterns that produce preferable predictions under weak supervision. Rather than relying solely on proxy classification losses, we design a pairwise ranking reward that serves as an in-batch proxy for ranking. For each sample, the reward reflects the fraction of correctly ordered positive-negative pairs, with a multiplicative bonus ($\gamma=2$) applied to clinically significant (csPCa, GG $\geq$ 3) cases to explicitly prioritize high-risk tumor detection.
For a mini-batch with positive set $\mathcal{P}$ and negative set $\mathcal{N}$, we compute a pairwise reward based on the ordering of classification scores:
\begin{equation}
r_i =
\begin{cases}
\frac{2}{|\mathcal{N}|}\sum_{j\in\mathcal{N}} \mathbf{1}[p_i > p_j] - 1, & y_i=1, \\
\frac{2}{|\mathcal{P}|}\sum_{j\in\mathcal{P}} \mathbf{1}[p_j > p_i] - 1, & y_i=0,
\end{cases}
\end{equation}
where $p_i$ is the predicted cancer probability. This reward promotes attention configurations that improve separation between malignant and benign cores. Clinically significant cancers receive amplified reward to prioritize high-grade disease.

To optimize this reward, we adopt a group-relative policy optimization strategy inspired by Group Relative Policy Optimization (GRPO) \cite{shao2024deepseekmath}, which computes advantages by comparing multiple rollouts of the same image. However, a challenge arises: our continuous spatial attention is mostly deterministic. Passing the same image through the policy network multiple times would yield identical attention maps, resulting in zero within-image reward variance and collapsing the policy gradients. To break this degeneracy and enable exploration, we inject continuous Gaussian noise directly into the attention logits prior to the softmax activation during the rollout phase: $\tilde{\mathbf{a}} = \mathbf{a} + \boldsymbol{\epsilon}$ with $\boldsymbol{\epsilon} \sim \mathcal{N}(0,\sigma^2)$, where $\sigma$ regulates the exploration scale. This stochastic injection generates $K$ different spatial configurations per forward pass, allowing the model to observe how subtle shifts in focal attention influence downstream classification rankings.

Finally, applying standard GRPO is suboptimal in our setting due to dataset class imbalance (86\% benign) and severe difficulty heterogeneity (e.g., small 10\% \emph{vs.} large 80\% tumor involvements). Standard rollout normalization treats all samples equally, allowing trivial benign cases to overwhelm the gradient signal. We address this by adopting Domain-Robust Policy Optimization (DRPO) \cite{dai2025qoq}. DRPO performs decoupled reward normalization and clusters samples hierarchically based on their clinical domain (cancer \emph{vs.} benign) and empirical difficulty (via K-Means clustering on the rollout reward vectors). By applying hierarchical temperature scaling to the computed advantages, DRPO dynamically upweights the gradients of rare cancer cases and hard-to-classify borderline lesions. During RL fine-tuning, we jointly optimize the DRPO policy loss alongside the supervised objectives, ensuring the model maintains core detection stability while explicitly nudging the policy toward optimally ranked predictions.

\subsection{Experiments} \label{sub:exp}
We compare against MedSAM-UNETR~\cite{alzate2023sam}, MedSAM-FT~\cite{MedSAM}, MicroSegNet~\cite{jiang2024microsegnet}, ProstNFound~\cite{wilson2024prostnfound}, and ProstNFound+~\cite{wilson2025prostnfoundprospectivestudyusing}, and conduct ablation studies on each component. Evaluation follows patient-level five-fold cross-validation with center-stratified splits; core-level scores are computed as mean heatmap activation within the needle–prostate intersection, and image-level csPCa scores from the classification head. We report AUROC and sensitivity at fixed specificities. Supervised training runs for 35 epochs (batch size 8, AdamW, lr$=2{\times}10^{-5}$, encoder lr$=1{\times}10^{-5}$, cosine annealing, weight decay $10^{-3}$, $\alpha{=}\beta{=}1$). RL fine-tuning runs for 35 additional epochs (batch size 16, $K{=}4$ rollouts, $\sigma{=}0.15$). Model selection uses validation AUC on high-involvement (${\geq}40\%$) cores.



\begin{table*}[t]
\centering
\caption{Prostate cancer detection performance in micro-ultrasound. Results are Mean $\pm$ Standard Deviation across five folds.}
\label{tab:merged_results}
\resizebox{\textwidth}{!}{
\begin{tabular}{l|ccccc}
\toprule
\multicolumn{6}{c}{\textbf{Auxiliary Image-Level Classification Head}} \\
\midrule
\multirow{2}{*}{\textbf{Method}} 
& \multicolumn{2}{c}{\textbf{All Cores}} 
& \multicolumn{3}{c}{\textbf{csPCa \emph{vs.} non-csPCa}} \\
\cmidrule(lr){2-3} \cmidrule(lr){4-6}
& \textbf{AUROC} & \textbf{Sens@80} 
& \textbf{AUROC} & \textbf{Sens@60} & \textbf{Sens@80} \\
\midrule
ProstNFound+ \cite{wilson2025prostnfoundprospectivestudyusing} 
& 75.4 $\pm$ 4.1 
& 56.3 $\pm$ 7.3 
& 78.5 $\pm$ 5.3 
& 81.8 $\pm$ 8.3 
& 58.2 $\pm$ 10.6 \\

\textbf{Prost-RL (Ours)} 
& \textbf{76.6 $\pm$ 3.4} 
& \textbf{58.0 $\pm$ 4.6} 
& \textbf{79.3 $\pm$ 5.8} 
& \textbf{83.1 $\pm$ 8.4} 
& \textbf{62.8 $\pm$ 12.6} \\
\midrule
\multicolumn{6}{c}{\textbf{Heatmap Decoder (Core-Level Detection)}} \\
\midrule
\multirow{2}{*}{\textbf{Method}} 
& \multicolumn{2}{c}{\textbf{All Cores}} 
& \multicolumn{2}{c}{\textbf{High Involvement}} 
& \textbf{csPCa} \\
\cmidrule(lr){2-3} \cmidrule(lr){4-5} \cmidrule(lr){6-6}
& \textbf{AUROC} & \textbf{Sens@80} 
& \textbf{AUROC} & \textbf{Sens@80} 
& \textbf{AUROC} \\
\midrule
MedSAM-UNETR \cite{10292632} 
& 71.4 $\pm$ 2.7 
& 50.7 $\pm$ 5.0 
& 79.4 $\pm$ 1.6 
& 63.9 $\pm$ 5.7 
& 71.5 $\pm$ 4.6 \\

MedSAM-FT \cite{ma2024segment} 
& 71.6 $\pm$ 2.0 
& 50.9 $\pm$ 4.2 
& 78.4 $\pm$ 0.9 
& 62.7 $\pm$ 1.9 
& 71.3 $\pm$ 3.8 \\

MicroSegNet \cite{jiang2024microsegnet} 
& 73.5 $\pm$ 2.2 
& 52.7 $\pm$ 3.2 
& 80.0 $\pm$ 0.6 
& 62.6 $\pm$ 1.7 
& 74.3 $\pm$ 5.2 \\

ProstNFound \cite{wilson2024prostnfound} 
& 76.8 $\pm$ 3.8 
& 61.2 $\pm$ 6.9 
& 82.0 $\pm$ 3.2 
& 70.7 $\pm$ 7.8 
& 78.0 $\pm$ 6.5 \\

ProstNFound+ \cite{wilson2025prostnfoundprospectivestudyusing} 
& 76.9 $\pm$ 3.5 
& 60.1 $\pm$ 5.6 
& 83.6 $\pm$ 2.4 
& 72.5 $\pm$ 4.9 
& 79.6 $\pm$ 4.1 \\
\midrule
\textbf{Prost-RL (Ours)} 
& \textbf{79.0 $\pm$ 3.5} 
& \textbf{64.6 $\pm$ 6.3} 
& \textbf{84.9 $\pm$ 2.5} 
& \textbf{76.3 $\pm$ 5.9} 
& \textbf{79.9 $\pm$ 5.5} \\
\bottomrule
\end{tabular}
}
\end{table*}

\begin{figure}[!t]
\includegraphics[width=\textwidth]{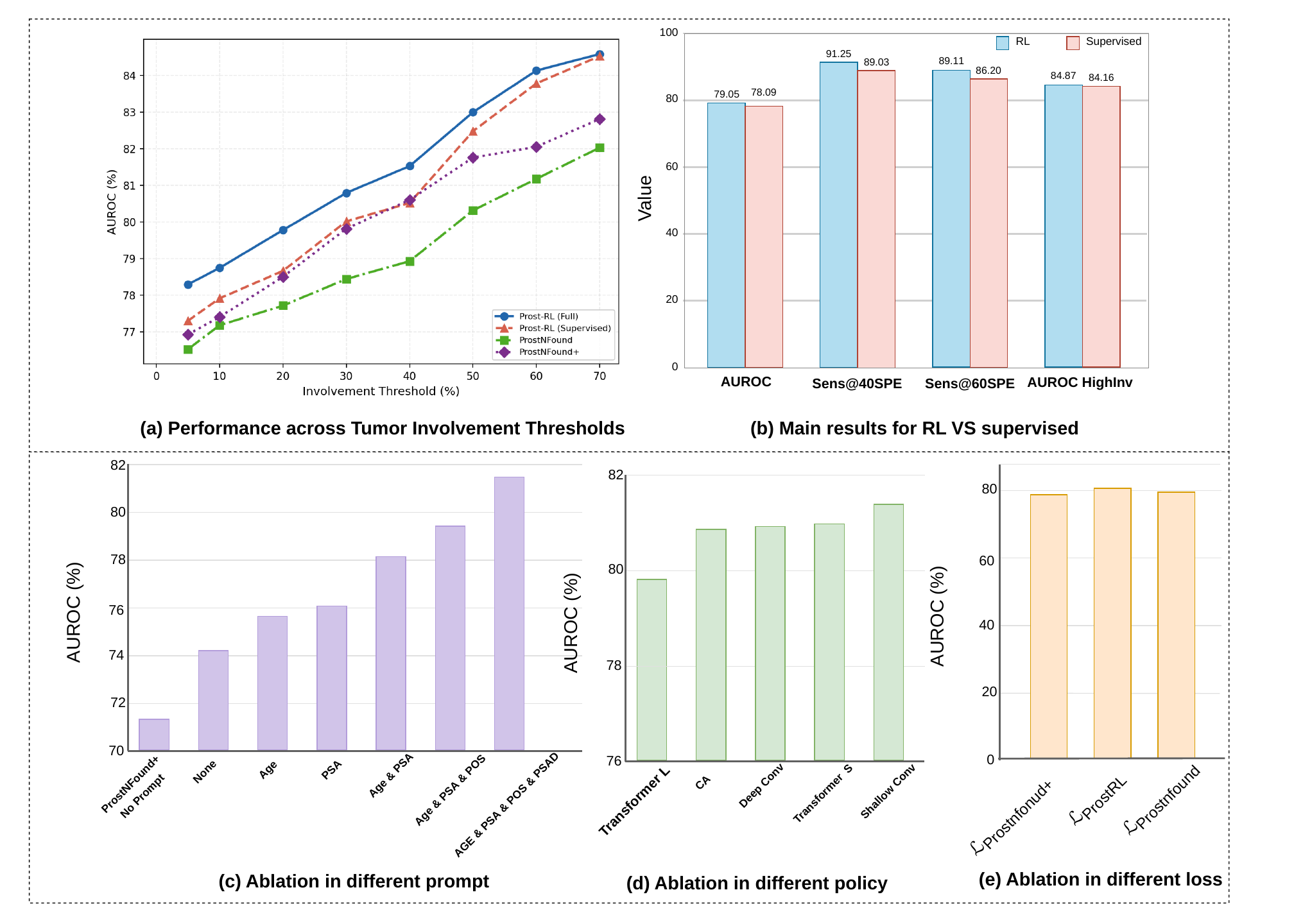}
\caption{Top: comparison of Prost-RL (Full) vs.\ supervised-only variant and other baselines. Bottom: ablation study results for different setting choices.}
\label{fig2}
\end{figure}

\section{Results and Discussion}
Results are summarized in Table~\ref{tab:merged_results}. Prost-RL consistently outperforms all state-of-the-art (SOTA) baselines across both evaluation heads. For core-level detection, it achieves $79.0 \pm 3.5$ AUROC, surpassing ProstNFound+ by $+2.1\%$, with sensitivity at 80\% specificity improving from $60.1\%$ to $64.6\%$. Gains extend to high-involvement cores ($84.9$ vs.\ $83.6$) and csPCa detection ($79.9$ vs.\ $79.6$). MedSAM-based approaches without explicit spatial guidance lag considerably behind, underscoring the benefit of learning where to look before decoding. The classification head mirrors this trend: Prost-RL achieves $79.3 \pm 5.8$ AUROC for csPCa vs.\ non-csPCa ($78.5 \pm 5.3$ for ProstNFound+), with sensitivity at 80\% specificity improving from $58.2\%$ to $62.8\%$, confirming that the shared attention policy consistently benefits both localization and global risk grading.

Fig.~\ref{fig2} part (a) plots mean core-level AUROC as a function of minimum cancer involvement threshold, isolating performance on progressively higher-confidence positive cores. All methods improve with higher involvement, consistent with larger tumors being more visually conspicuous. Parts (a) and (b), illustrate the contribution of our two main additions: the supervised stage alone (Prost-RL Supervised, incorporating the attention policy and noise-robust losses) surpasses both baselines, ProstNFound and ProstNFound+, at every threshold, demonstrating that explicit spatial attention modulation and SCE meaningfully improve localization. The full Prost-RL model widens this gap further across all thresholds, with the margin over the supervised variant growing at lower involvement levels, precisely the regime where weak, noisy labels are most problematic, suggesting that APO does not merely boost overall confidence but specifically improves the model's ability to rank and localize subtle, low-involvement lesions that supervised training alone struggles to distinguish.

\begin{figure}[t]
\includegraphics[width=\textwidth]{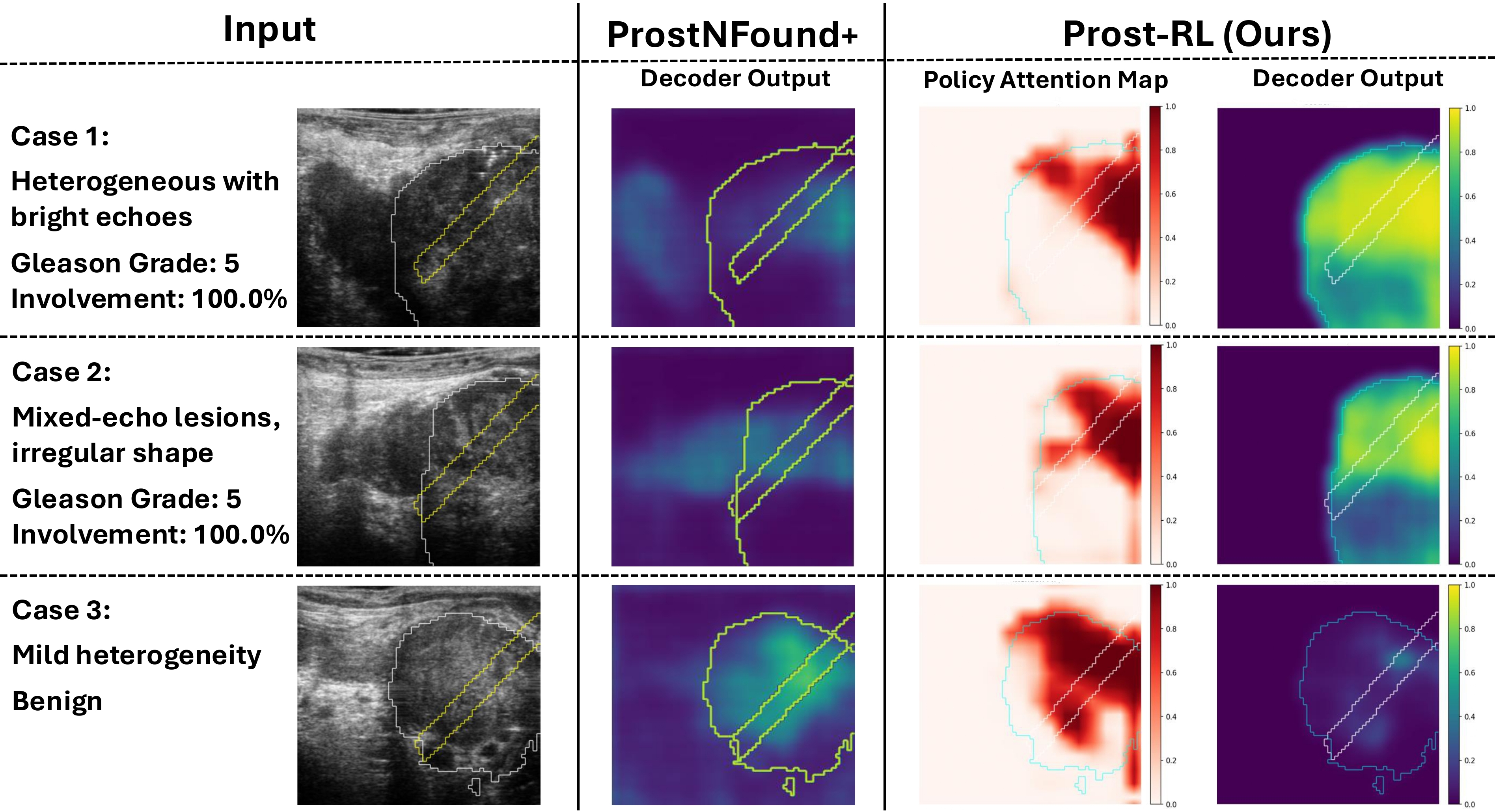}
\caption{Example heatmaps generated by ProstRL and ProstNFound+, shown alongside ground‑truth labels and radiologist annotations.} \label{fig3}
\end{figure}

\noindent\textbf{Ablation Studies and Qualitative Evaluation.}
Fig.~\ref{fig2} (part c--e) ablates the key design choices of Prost-RL. \textbf{Prompt ablation (c):} even without any clinical prompts, Prost-RL surpasses ProstNFound+ by a clear margin, confirming that the attention policy and noise-robust training provide independent gains. Performance improves monotonically as prompts are added, with the full set yielding the best results, consistent with each feature providing complementary clinical context. \textbf{Policy architecture (d):} all tested architectures (transformer and convolutional variants) perform comparably, indicating that the RL training strategy drives the gains rather than a specific network design. \textbf{Loss function (e):} our proposed loss outperforms the needle-region cross-entropy of ProstNFound and the MIL proportion BCE of ProstNFound+, demonstrating that explicit noise robustness via SCE and negative-entropy regularization is beneficial in this weakly supervised, label-noisy setting. Fig.~\ref{fig3} compares outputs qualitatively. For malignant cases, Prost-RL produces sharper, more spatially focused heatmaps confined to biopsy-confirmed regions, while ProstNFound+ yields diffuse activations. The policy attention map provides an interpretable spatial prior correlated with the model's decision, supporting clinician trust. For the benign case, Prost-RL correctly suppresses needle-region activation where ProstNFound+ does not.

Prost-RL's gains arise from explicit spatial attention guiding decoding under weak supervision, SCE with entropy regularization enforcing sharp heatmaps, and APO refining the policy toward better malignant-benign separation under class imbalance. Extension to 3D volumes remains as an important future works.

\begin{credits}
\subsubsection{\ackname}
This work was supported in part by the Canadian Institutes of Health Research (CIHR), the Natural Sciences and Engineering Research Council of Canada (NSERC), Vector Institute, and through computational resources and services provided by Advanced Research Computing at the University of British Columbia. P. Mousavi is supported in part by Canada CIFAR AI Chair and Canada Research Chair.
\end{credits}

\bibliographystyle{splncs04}
\bibliography{refs}

\end{document}